\documentclass{article} 
\usepackage{nips14submit_e,times}
\usepackage{hyperref}
\usepackage{url}

\usepackage{amsmath,amsfonts,amsbsy}
\usepackage{latexsym}
\usepackage{graphicx}
\usepackage{algorithm,algorithmic}
\usepackage{multicol,multirow}
\usepackage{subfigure}
\usepackage{url}
\usepackage{color}

\newcommand*\samethanks[1][\value{footnote}]{\footnotemark[#1]}

\title{{PVANET}: Deep but Lightweight Neural Networks for Real-time Object Detection}

\author{
Kye-Hyeon Kim\thanks{These authors contributed equally. Corresponding author: Sanghoon Hong}, Sanghoon Hong\samethanks, Byungseok Roh\samethanks, Yeongjae Cheon, and Minje Park\\
Intel Imaging and Camera Technology\\
21 Teheran-ro 52-gil, Gangnam-gu, Seoul 06212, Korea\\
\texttt{\{kye-hyeon.kim, sanghoon.hong, peter.roh,}\\
\texttt{yeongjae.cheon, minje.park\}@intel.com} \\
}

\nipsfinalcopy 

\begin{document}

\maketitle

\begin{abstract}
This paper presents how we can achieve the state-of-the-art accuracy in multi-category object detection task while minimizing the computational cost by adapting and combining recent technical innovations. Following the common pipeline of ``CNN feature extraction + region proposal + RoI classification'', we mainly redesign the feature extraction part, since region proposal part is not computationally expensive and classification part can be efficiently compressed with common techniques like truncated SVD. Our design principle is {\em ``less channels with more layers''} and adoption of some building blocks including concatenated ReLU, Inception, and HyperNet. The designed network is deep and thin and trained with the help of batch normalization, residual connections, and learning rate scheduling based on plateau detection.
We obtained solid results on well-known object detection benchmarks: 83.8\% mAP (mean average precision) on VOC2007 and 82.5\% mAP on VOC2012 (2nd place), while taking only 750ms/image on Intel i7-6700K CPU with a single core and 46ms/image on NVIDIA Titan X GPU. Theoretically, our network requires only 12.3\% of the computational cost compared to ResNet-101, the winner on VOC2012.
\end{abstract}

\section{Introduction}

Convolutional neural networks (CNNs) have made impressive improvements in object detection for several years.
Thanks to many innovative work, recent object detection systems have met acceptable accuracies for commercialization in a broad range of markets like automotive and surveillance. In terms of detection speed, however, even the best algorithms are still suffering from heavy computational cost. Although recent work on network compression and quantization shows promising result, it is important to reduce the computational cost in the network design stage.

This paper presents our lightweight feature extraction network architecture for object detection, named {\sc PVANET}\footnote{The code and the trained models are available at \url{https://github.com/sanghoon/pva-faster-rcnn}}, which achieves real-time object detection performance without losing accuracy compared to the other state-of-the-art systems:

\begin{itemize}
\item Computational cost: 7.9GMAC for feature extraction with 1065x640 input (cf. ResNet-101 \cite{HeK2016cvpr}: 80.5GMAC\footnote{ResNet-101 used multi-scale testing without mentioning additional computation cost. If we take this into account, ours requires only $<$7\% of the computational cost compared to ResNet-101.})
\item Runtime performance: 750ms/image (1.3FPS) on Intel i7-6700K CPU with a single core; 46ms/image (21.7FPS) on NVIDIA Titan X GPU
\item Accuracy: 83.8\% mAP on VOC-2007; 82.5\% mAP on VOC-2012 (2nd place)
\end{itemize}

The key design principle is ``less channels with more layers''. Additionally, our networks adopted some recent building blocks while some of them have not been verified their effectiveness on object detection tasks:

\begin{itemize}
\item Concatenated rectified linear unit (C.ReLU) \cite{ShangW2016icml} is applied to the {\em early stage} of our CNNs (i.e., first several layers from the network input) to reduce the number of computations by half without losing accuracy.
\item Inception \cite{SzegedyC2015cvpr} is applied to the remaining of our feature generation sub-network.
An Inception module produces output activations of different sizes of receptive fields, so that increases the variety of receptive field sizes in the previous layer. We observed that stacking up Inception modules can capture widely varying-sized objects more effectively than a linear chain of convolutions.
\item We adopted the idea of multi-scale representation like HyperNet \cite{KongT2016cvpr} that combines several intermediate outputs so that multiple levels of details and non-linearities can be considered simultaneously.
\end{itemize}

We will show that our thin but deep network can be trained effectively with batch normalization \cite{SzegedyC2015icml}, residual connections \cite{HeK2016cvpr}, and learning rate scheduling based on plateau detection \cite{HeK2016cvpr}.

In the remaining of the paper, we describe our network design briefly (Section \ref{sec:design}) and summarize the detailed structure of {\sc PVANET} (Section \ref{sec:structure}). Finally we provide some experimental results on VOC-2007 and VOC-2012 benchmarks, with detailed settings for training and testing (Section \ref{sec:experiments}).

\section{Details on Network Design}
\label{sec:design}

\subsection{C.ReLU: Earlier building blocks in feature generation}
\label{subsec:c_relu}

C.ReLU is motivated from an interesting observation of intermediate activation patterns in CNNs.
In the early stage, output nodes tend to be ``paired'' such that one node's activation is the opposite side of another's.
From this observation, C.ReLU reduces the number of output channels by half, and doubles it by simply concatenating the same outputs {\em with negation}, which leads to 2x speed-up of the early stage without losing accuracy.

\begin{figure}[t]
\centerline{
\includegraphics[height=0.12\paperheight]{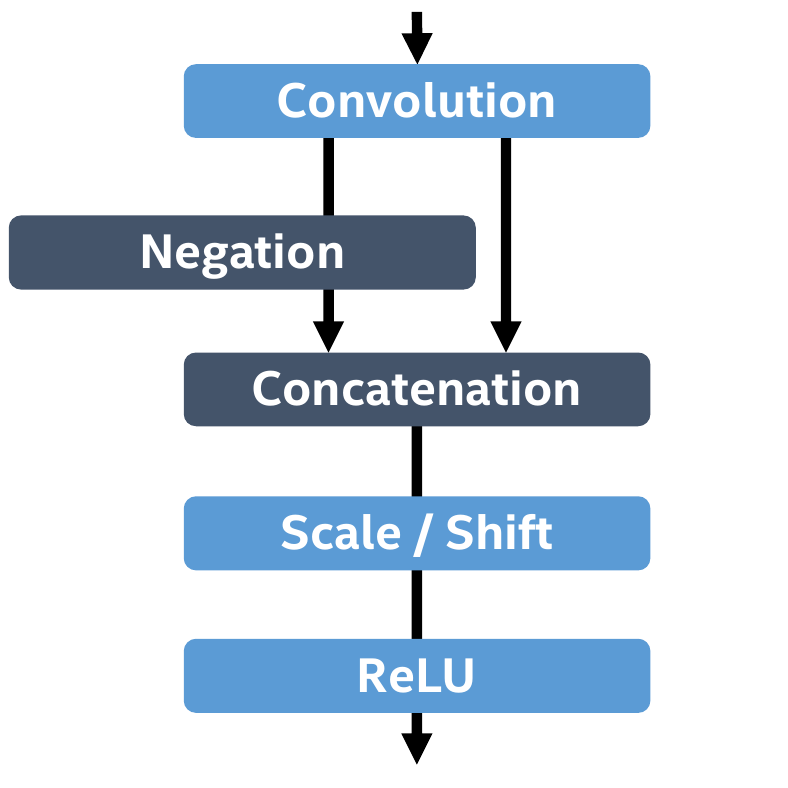}
}
\caption{Our C.ReLU building block. {\bf Negation} simply multiplies $-1$ to the output of Convolution. {\bf Scale / Shift} applies trainable weight and bias to each channel, allowing activations in the negated part to be adaptive.}
\label{fig:block_c_relu}
\end{figure}

Figure \ref{fig:block_c_relu} illustrates our C.ReLU implementation.
Compared to the original C.ReLU, we append scaling and shifting after concatenation to allow that each channel's slope and activation threshold can be different from those of its opposite channel.

\subsection{Inception: Remaining building blocks in feature generation}
\label{subsec:inception}

For object detection tasks, Inception has neither been widely applied to existing work, nor been verified its effectiveness.
We found that Inception can be one of the most cost-effective building block for capturing both small and large objects in an input image. To Learn visual patterns for capturing large object, output features of CNNs should correspond to sufficiently large receptive fields, which can be easily fulfilled by stacking up convolutions of 3x3 or larger kernels.
On the other hand, for capturing small-sized objects, output features should correspond to sufficiently small receptive fields to localize small regions of interest precisely.

\begin{figure}[t]
\centerline{
\includegraphics[width=1.0\columnwidth]{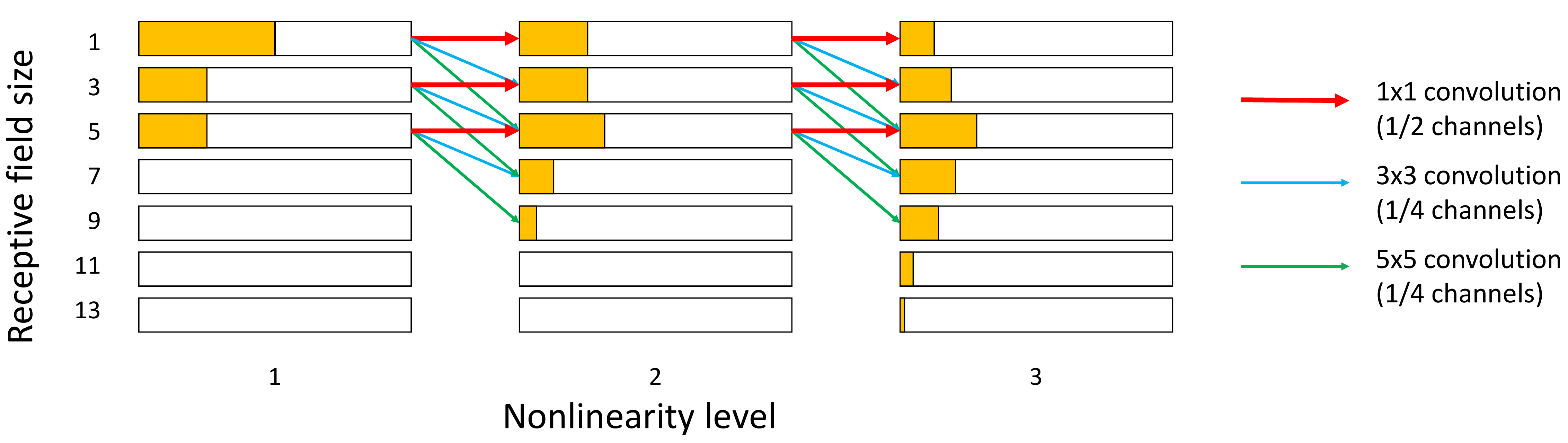}
}
\caption{Example of a distribution of (expected) receptive field sizes of intermediate outputs in a chain of 3 Inception modules.
Each module concatenates 3 convolutional layers of different kernel sizes, 1x1, 3x3 and 5x5, respectively. The number of output channels in each module is set to $\{ 1/2, 1/4, 1/4 \}$ of the number of channels from the previous module, respectively. A latter Inception module can learn visual patterns of wider range of sizes, as well as having higher level of nonlinearity.
}
\label{fig:inception_distribution_example}
\end{figure}

\begin{figure}[t]
\centerline{
\includegraphics[height=0.14\paperheight]{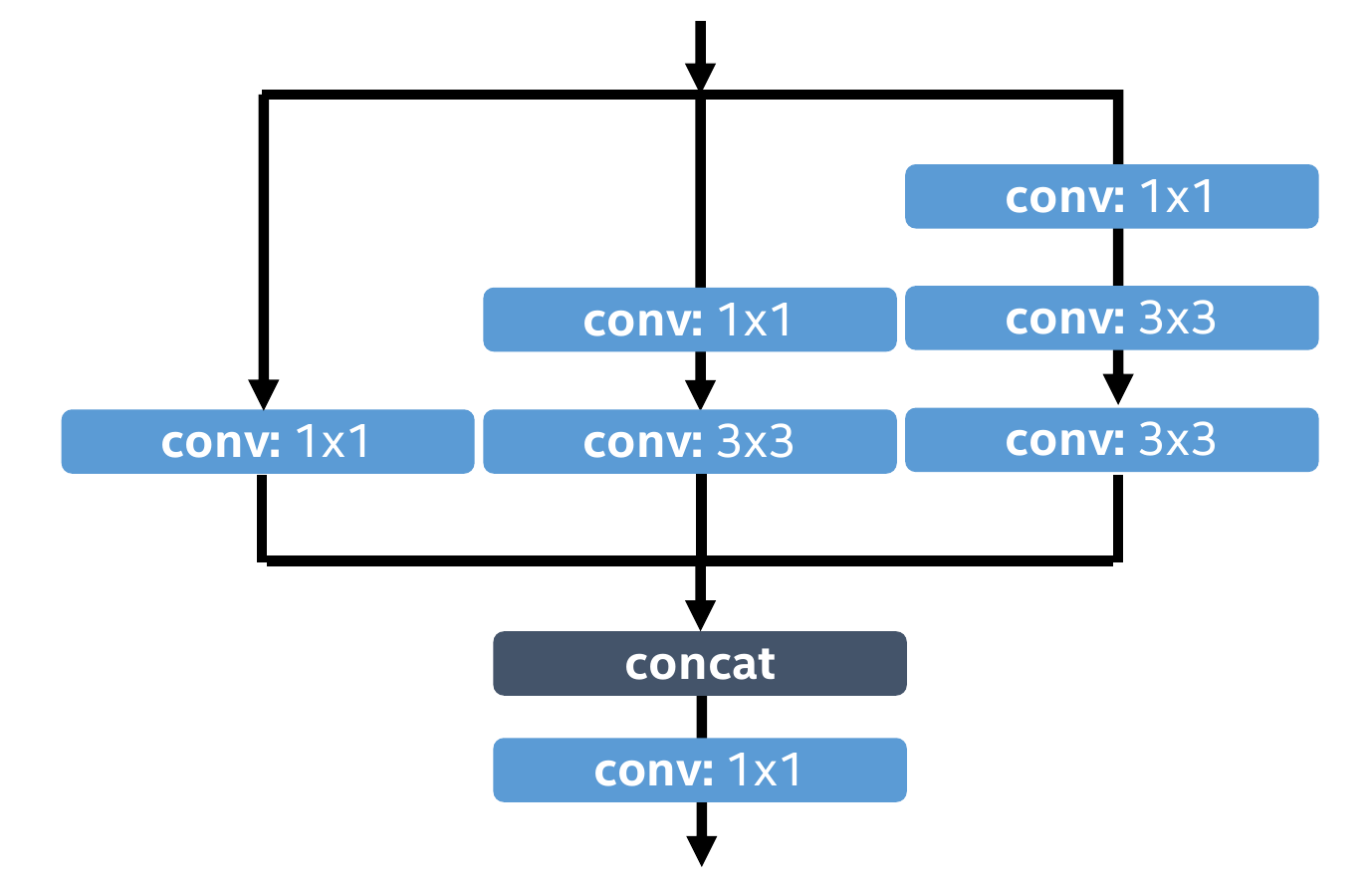}
\includegraphics[height=0.14\paperheight]{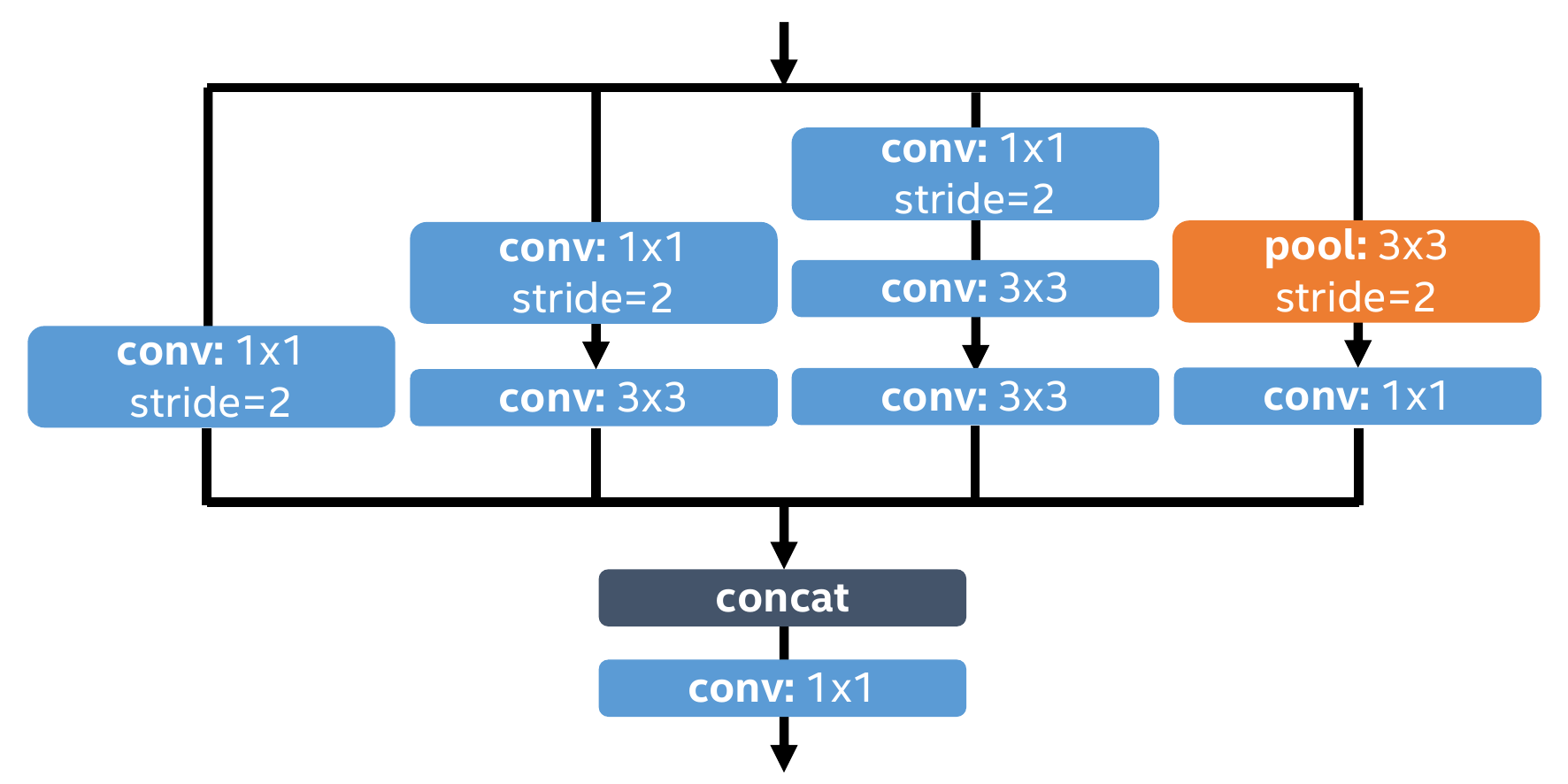}
}
\caption{(Left) Our Inception building block. 5x5 convolution is replaced with two 3x3 convolutional layers for efficiency. (Right) Inception for reducing feature map size by half.}
\label{fig:block_inception}
\end{figure}

Figure \ref{fig:inception_distribution_example} clearly shows that Inception can fulfill both requirements. 1x1 convolution plays the key role to this end, by {\em preserving the receptive field} of the previous layer. Just increasing the nonlinearity of input patterns, it slows down the growth of receptive fields for some output features so that small-sized objects can be captured precisely.
Figure \ref{fig:block_inception} illustrates our Inception implementation. 5x5 convolution is replaced with a sequence of two 3x3 convolutions.

\subsection{HyperNet: Concatenation of multi-scale intermediate outputs}
\label{subsec:hypernet}

Multi-scale representation and its combination are proven to be effective in many recent deep learning tasks \cite{KongT2016cvpr,BellS2016cvpr,HariharanB2015cvpr}.
Combining fine-grained details with highly-abstracted information in feature extraction layer helps the following region proposal network and classification network to detect objects of different scales.
However, since the direct concatenation of all abstraction layers may produce redundant information with much higher compute requirement we need to design the number of different abstraction layers and the layer numbers of abstraction carefully.
If you choose the layers which are too early for object proposal and classification, it would be little help when we consider additional compute complexity.

Our design choice is not different from the observation from ION \cite{BellS2016cvpr} and HyperNet \cite{KongT2016cvpr}, 
which combines 1) the last layer and 2) two intermediate layers whose scales are 2x and 4x of the last layer, respectively.
We choose the middle-sized layer as a reference scale (= 2x), and concatenate the 4x-scaled layer and the last layer with down-scaling (pooling) and up-scaling (linear interpolation), respectively.

\subsection{Deep network training}

It is widely accepted that as network goes deeper and deeper, the training of network becomes more troublesome. We solve this issue by adopting residual structures \cite{HeK2016cvpr}. Unlike the original residual training idea, we add residual connections onto inception layers as well to stabilize the later part of our deep network architecture.

We also add Batch normalization \cite{SzegedyC2015icml} layers before all ReLU activation layers.
Mini-batch sample statistics are used during pre-training, and moving-averaged statistics are used afterwards as fixed scale-and-shift parameters.

Learning rate policy is also important to train network successfully.
Our policy is to control the learning rate dynamically, based on plateau detection \cite{HeK2016cvpr}.
We measure the moving average of loss, and decide it to be {\em on-plateau} if its improvement is below a threshold during a certain period of iterations.
Whenever the plateau is detected, the learning rate is decreased by a constant factor.
In experiments, our learning rate policy gave a significant gain of accuracy.




\section{Faster R-CNN with our feature extraction network}
\label{sec:structure}

\begin{table}
\hspace{-1.3cm}
\scriptsize
\begin{tabular}{c|c|c|c|c|c|ccccc|r|r}
Name & Type & Stride & Output & Residual     & C.ReLU        & \multicolumn{5}{|c|}{Inception}         & \# params & MAC\\
     &      &        & size   &              & \#1x1-KxK-1x1 & \#1x1 & \#3x3 & \#5x5 & \#pool & \#out  &           & \\
\hline
conv1\_1 & 7x7 C.ReLU    & 2 & 528x320x32 &    & X-16-X    &    & & & & & 2.4K & 397M\\
pool1\_1 & 3x3 max-pool & 2 & 264x160x32 &    &           &    & & & & &      &   \\
conv2\_1 & 3x3 C.ReLU &   & 264x160x64 & O  & 24-24-64  &    & & & & & 11K & 468M\\
conv2\_2 & 3x3 C.ReLU &   & 264x160x64 & O  & 24-24-64  &    & & & & & 9.8K & 414M\\
conv2\_3 & 3x3 C.ReLU &   & 264x160x64 & O  & 24-24-64  &    & & & & & 9.8K & 414M\\
conv3\_1 & 3x3 C.ReLU & 2 & 132x80x128 & O  & 48-48-128 &    & & & & & 44K & 468M\\
conv3\_2 & 3x3 C.ReLU &   & 132x80x128 & O  & 48-48-128 &    & & & & & 39K & 414M\\
conv3\_3 & 3x3 C.ReLU &   & 132x80x128 & O  & 48-48-128 &    & & & & & 39K & 414M\\
conv3\_4 & 3x3 C.ReLU &   & 132x80x128 & O  & 48-48-128 &    & & & & & 39K & 414M\\
conv4\_1 & Inception   & 2 & 66x40x256 & O  &           & 64 & 48-128 & 24-48-48 & 128 & 256 & 247K & 653M\\
conv4\_2 & Inception   &   & 66x40x256 & O  &           & 64 & 64-128 & 24-48-48 &     & 256 & 205K & 542M\\
conv4\_3 & Inception   &   & 66x40x256 & O  &           & 64 & 64-128 & 24-48-48 &     & 256 & 205K & 542M\\
conv4\_4 & Inception   &   & 66x40x256 & O  &           & 64 & 64-128 & 24-48-48 &     & 256 & 205K & 542M\\
conv5\_1 & Inception   & 2 & 33x20x384 & O  &           & 64 & 96-192 & 32-64-64 & 128 & 384 & 573K & 378M\\
conv5\_2 & Inception   &   & 33x20x384 & O  &           & 64 & 96-192 & 32-64-64 &     & 384 & 418K & 276M\\
conv5\_3 & Inception   &   & 33x20x384 & O  &           & 64 & 96-192 & 32-64-64 &     & 384 & 418K & 276M\\
conv5\_4 & Inception   &   & 33x20x384 & O  &           & 64 & 96-192 & 32-64-64 &     & 384 & 418K & 276M\\
\hline
downscale & 3x3 max-pool & 2 & 66x40x128 &    &           &    & & & & &      &   \\
upscale   & 4x4 deconv   & 2 & 66x40x384 &    &           &    & & & & & 6.2K & 16M\\
concat    & concat       &   & 66x40x768 &    &           &    & & & & &      &   \\
convf     & 1x1 conv     &   & 66x40x512 &    &           &    & & & & & 393K & 1038M\\
\hline
Total     &              &   &           &    &           &    & & & & & 3282K & 7942M\\
\end{tabular}
\caption{
The detailed structure of {\sc PVANET}.
All conv layers are combined with batch normalization, channel-wise scaling and shifting, and ReLU activation layers.
Theoretical computational cost is given as the number of adds and multiplications (MAC), assuming that the input image size is 1056x640.
{\bf KxK C.ReLU} refers to a sequence of ``1x1 - KxK - 1x1'' conv layers, where KxK is a C.ReLU block as in Figure \ref{fig:block_c_relu}. conv1\_1 has no 1x1 conv layer. ``C.ReLU'' column shows the number of output channels of each conv layer.
For {\bf Residual}, 1x1 conv is applied for projecting pool1\_1 into conv2\_1, conv2\_3 into conv3\_1, conv3\_4 into conv4\_1, and conv4\_4 into conv5\_1.
{\bf Inception} consists of four sub-sequences: 1x1 conv (\#1x1); ``1x1 - 3x3'' conv (\#3x3); ``1x1 - 3x3 - 3x3'' conv (\#5x5); ``3x3 max-pool - 1x1 conv'' (\#pool, only for stride 2). ``\#out'' refers to 1x1 conv after concatenating those sub-sequences. The number of output channels of each conv layer is shown.
{\bf Multi-scale features} are obtained by four steps: conv3\_4 is down-scaled into ``downscale'' by 3x3 max-pool with stride 2;
conv5\_4 is up-scaled into ``upscale'' by 4x4 channel-wise deconvolution whose weights are fixed as bilinear interpolation; ``downscale'', conv4\_4 and ``upscale'' are combined into ``concat'' by channel-wise concatenation; after 1x1 conv, the final output is obtained (convf).
}
\label{table:pvanet_structure}
\end{table}

Table \ref{table:pvanet_structure} shows the whole structure of {\sc PVANET}.
In the early stage (conv1\_1, ..., conv3\_4), C.ReLU is adapted to convolutional layers to reduce the computational cost of KxK conv by half. 1x1 conv layers are added before and after the KxK conv, in order to reduce the input size and then enlarge the representation capacity, respectively.



Three intermediate outputs from conv3\_4 (with down-scaling), conv4\_4, and conv5\_4 (with up-scaling) are combined into the 512-channel multi-scale output features (convf), which are fed into the Faster R-CNN modules:

\begin{itemize}
\item For computational efficiency, only the first 128 channels in convf are fed into the region proposal network (RPN). Our RPN is a sequence of ``3x3 conv (384 channels) - 1x1 conv (25x(2+4) = 150 channels\footnote{RPN produces 2 predicted scores (foreground and background) and 4 predicted values of the bounding box for each anchor. Our RPN uses 25 anchors of 5 scales (3, 6, 9, 16, 25) and 5 aspect ratios (0.5, 0.667, 1.0, 1.5, 2.0).})'' layers to generate regions of interest (RoIs) from 
\item R-CNN takes all 512 channels in convf. For each RoI, 6x6x512 tensor is generated by RoI pooling, and then passed through a sequence of fully-connected layers of ``4096 - 4096 - (21+84)'' output nodes.\footnote{For 20-class object detection, R-CNN produces 21 predicted scores (20 classes + 1 background) and 21x4 predicted values of 21 bounding boxes.}
\end{itemize}


\section{Experimental results}
\label{sec:experiments}

\subsection{Training and testing}

{\sc PVANET} was pre-trained with ILSVRC2012 training images for 1000-class image classification.\footnote{\url{http://www.image-net.org/challenges/LSVRC/2012/}}
All images were resized into 256x256, and 192x192 patches were randomly cropped and used as the network input.
The learning rate was initially set to 0.1, and then decreased by a factor of $1/\sqrt{10} \approx 0.3165$ whenever a plateau is detected. Pre-training terminated if the learning rate drops below $1e-4$, which usually requires about 2M iterations.

Then {\sc PVANET} was trained with the union set of MS COCO\footnote{\url{http://mscoco.org/dataset/}} trainval, VOC2007\footnote{\url{http://host.robots.ox.ac.uk/pascal/VOC/voc2007/}} trainval and VOC2012\footnote{\url{http://host.robots.ox.ac.uk/pascal/VOC/voc2012/}} trainval.
Fine-tuning with VOC2007 trainval and VOC2012 trainval was also required afterwards, since the class definitions in MS COCO and VOC competitions are slightly different.
Training images were resized randomly such that a shorter edge of an image to be between 416 and 864.

For PASCAL VOC evaluations, each input image was resized such that its shorter edge to be 640.
All parameters related to Faster R-CNN were set as in the original work \cite{RenS2015nips} except for the number of proposal boxes before non-maximum suppression (NMS) ($= 12000$) and the NMS threshold ($= 0.4$). All evaluations were done on Intel i7-6700K CPU with a single core and NVIDIA Titan X GPU.

\subsection{VOC2007}

Table \ref{table:voc2007} shows the accuracy of our models in different configurations.\footnote{On Sep. 19, 2016, we updated the mAP numbers according to the latest version of the evaluation code in py-faster-rcnn.} Thanks to Inception (Section \ref{subsec:inception}) and multi-scale features (Section \ref{subsec:hypernet}), our RPN generated initial proposals very accurately.
Since the results imply that more than 200 proposals does not give notable benefits to detection accuracy, we fixed the number of proposals to 200 in the remaining experiments.
We also measured the performance with bounding-box voting \cite{GidarisS2015iccv}, while iterative regression was not applied.

\begin{table}
\centering
\begin{tabular}{l|r|r|r|r|r}
Model         & Proposals & Recall (\%) & mAP (\%)   & Time (ms) & FPS\\
\hline
{\sc PVANET}  & 300       & 98.9        & 83.6       & 48.5      & 20.6\\
              & 200       & 98.3        & 83.5       & 42.2      & 23.7\\
              & 100       & 97.0        & 83.2       & 40.0      & 25.0\\
              & 50        & 94.7        & 82.1       & 26.8      & 37.3\\
\hline
{\sc PVANET+} & 200       & 98.3        & {\bf 83.8} & 46.1      & 21.7\\
{\sc PVANET+} (compressed) & 200 & 98.3 & {\bf 82.9} & 31.9      & 31.3\\
\end{tabular}
\caption{Performance on VOC2007-test benchmark data. ``Recall'' refers to a ratio of ``true positive (TP)'' boxes among the proposals, considering a box as TP if the intersection-over-union (IoU) score with its maximally-overlapped ground-truth box is $\geq 0.5$. {\sc PVANET+} denotes that bounding-box voting is applied, and {\sc PVANET+} (compressed) denotes that fully-connected layers in R-CNN are compressed.
}
\label{table:voc2007}
\end{table}

Faster R-CNN consists of fully-connected layers, which can be compressed easily without a significant drop of accuracy \cite{GirshickR2015iccv}. We compressed the fully-connected layers of ``4096 - 4096'' into to ``512 - 4096 - 512 - 4096'' by the truncated singular value decomposition (SVD), with some fine-tuning after that. The compressed network achieved 82.9\% mAP (-0.9\%) and ran in 31.3 FPS (+9.6 FPS).

\subsection{VOC2012}

Table \ref{table:voc2012} summarizes comparisons between {\sc PVANET+} and some state-of-the-art networks \cite{HeK2016cvpr,RenS2015nips,Jif2016nips} from the PASCAL VOC2012 leaderboard.\footnote{\url{http://host.robots.ox.ac.uk:8080/leaderboard/displaylb.php?challengeid=11&compid=4}}

\begin{table}
\small
\centering
\begin{tabular}{l|r|r|r|r|r|r|c}
Model                     & \multicolumn{4}{|c|}{Computation cost (MAC)} & \multicolumn{2}{|c|}{Running time} & mAP\\
                          & Shared CNN & RPN  & Classifier & Total       & ms   & x({\sc PVANET})     & (\%)\\
\hline
{\sc PVANET+}             & 7.9        & 1.3  & 27.7       & 37.0        & 46   & 1.0                 & 82.5\\
\hline
Faster R-CNN + ResNet-101 & 80.5       & N/A  & 219.6      & 300.1       & 2240 & 48.6                & 83.8\\
Faster R-CNN + VGG-16     & 183.2      & 5.5  & 27.7       & 216.4       & 110  & 2.4                 & 75.9\\
R-FCN + ResNet-101        & 122.9      & 0    & 0          & 122.9       & 133  & 2.9                 & 82.0\\
\end{tabular}
\caption{Comparisons between our network and some state-of-the-arts in the PASCAL VOC2012 leaderboard.
{\sc PVANET+} denotes {\sc PVANET} with bounding-box voting.
We assume that {\sc PVANET} takes a 1056x640 image and the number of proposals is 200.
Competitors' MAC are estimated from their Caffe prototxt files which are publicly available. All testing-time configurations are the same with the original articles \cite{HeK2016cvpr,Jif2016nips,RenS2015nips}.
Competitors' runtime performances are also therein, while we projected the original values with assuming that NVIDIA Titan X is 1.5x faster than NVIDIA K40.
}
\label{table:voc2012}
\end{table}

Our {\sc PVANET+} achieved 82.5\% mAP, the 2nd place on the leaderboard, outperforming all other competitors except for ``Faster R-CNN + ResNet-101''.
However, the top-performer uses ResNet-101 which is much heavier than {\sc PVANET}, as well as several time-consuming techniques such as global contexts and multi-scale testing, leading to 40x (or more) slower than ours.
In Table \ref{table:voc2012}, we also compare mAP with respect to the computational cost. Among the networks performing over 80\% mAP, {\sc PVANET+} is the only network running $\leq 50$ms. Taking its accuracy and computational cost into account, our {\sc PVANET+} is the most efficient network in the leaderboard.

\section{Conclusions}

In this paper, we showed that the current networks are highly redundant and we can design a thin and light network which is capable enough for complex vision tasks.
Elaborate adoption and combination of recent technical innovations on deep learning makes us possible to re-design the feature extraction part of the Faster R-CNN framework to maximize the computational efficiency.
Even though the proposed network is designed for object detection, we believe our design principle can be widely applicable to other tasks like face recognition and semantic analysis.

Our network design is completely independent of network compression and quantization.
All kinds of recent compression and quantization techniques are applicable to our network as well to further increase the actual performance in real applications.
As an example, we showed that a simple technique like truncated SVD could achieve a notable improvement in the runtime performance based on our network.

\renewcommand\refname{\subsubsection*{References}}
\small{
\bibliographystyle{unsrt}
\bibliography{khk}
}

\end{document}